\documentclass{article}
\usepackage{algorithm} 
\usepackage[numbers,sort&compress]{natbib}
\usepackage{algpseudocode} 
\usepackage{booktabs}
\usepackage{PRIMEarxiv}
\usepackage[table,xcdraw]{xcolor}
\usepackage[utf8]{inputenc} 
\usepackage[T1]{fontenc}    
\usepackage{hyperref}       
\usepackage{url}            
\usepackage{booktabs}       
\usepackage{amsfonts}       
\usepackage{nicefrac}       
\usepackage{microtype}      
\usepackage{lipsum}
\usepackage{fancyhdr}       
\usepackage{graphicx}       
\graphicspath{{media/}}     
\usepackage{adjustbox}
\usepackage{algorithm} 
\usepackage{algpseudocode}
\usepackage[section]{placeins}
\usepackage{amsmath}
\usepackage{amssymb}
\usepackage{booktabs}
\usepackage{multirow}
\usepackage{multirow,diagbox,tabularx,blindtext}
\usepackage{threeparttable}
\usepackage{color}
\usepackage{float}
\usepackage{tabularray}
\usepackage{longtable}
\usepackage[numbers,sort&compress]{natbib}

\usepackage{lineno}
\nolinenumbers
\title{Structured Prompting and Multi-Agent Knowledge Distillation for Traffic Video Interpretation and Risk Inference}

\author{
 Yunxiang Yang$^{*}$, Ningning Xu$^{*}$, Jidong J. Yang$^{\dagger}$ \\
Smart Mobility and Infrastructure Lab\\
College of Engineering, University of Georgia \\
  \texttt{\{yyang117, Ningning.Xu, Jidong.Yang\}@uga.edu} \\
}

\begin{document}
\maketitle

\renewcommand*{\thefootnote}{\fnsymbol{footnote}}
\footnotetext[1]{Equal contribution.}
\footnotetext[2]{Corresponding author.}

\begin{abstract}
Comprehensive highway scene understanding and robust traffic risk inference are vital for advancing Intelligent Transportation Systems (ITS) and autonomous driving. Traditional approaches often struggle with scalability and generalization, particularly under the complex and dynamic conditions of real-world environments. To address these challenges, we introduce a novel structured prompting and knowledge distillation framework that enables automatic generation of high-quality traffic scene annotations and contextual risk assessments.  Our framework orchestrates two large Vision-Language Models (VLMs): GPT-4o and o3-mini, using a structured Chain-of-Thought (CoT) strategy to produce rich, multi-perspective outputs. These outputs serve as knowledge-enriched pseudo-annotations for supervised fine-tuning of a much smaller student VLM. The resulting compact 3B-scale model, named VISTA (Vision for Intelligent Scene and Traffic Analysis), is capable of understanding low-resolution traffic videos and generating semantically faithful, risk-aware captions.  Despite its significantly reduced parameter count, VISTA achieves strong performance across established captioning metrics (BLEU-4, METEOR, ROUGE-L, and CIDEr) when benchmarked against its teacher models. This demonstrates that effective knowledge distillation and structured multi-agent supervision can empower lightweight VLMs to capture complex reasoning capabilities. The compact architecture of VISTA facilitates efficient deployment on edge devices, enabling real-time risk monitoring without requiring extensive infrastructure upgrades.
\end{abstract}


\section{Introduction}
The rapid advancement of Intelligent Transportation Systems (ITS) and autonomous driving technologies has created an urgent need for comprehensive road scene understanding and robust traffic risk inference \cite{Rivera2025, Zhang2024b, Wang2023}. These systems require real-time, reliable interpretation of complex, dynamic environments, where factors such as weather, pavement conditions, traffic congestion, and unexpected hazards interact in unpredictable ways. Numerous studies have shown that adverse environmental factors, such as rain, snow, fog, and poor visibility, substantially increase the likelihood and severity of traffic accidents \cite{Theofilatos2014}. Traditional approaches, largely dependent on manually annotated datasets and task-specific neural networks, struggle with scalability, adaptability, and generalization, particularly in real-world, high-variability conditions \cite{Rivera2025, Zhang2024b, Wang2023}. As transportation systems become more interconnected and data-rich, the limitations of the traditional methods underscore the necessity for flexible, generalizable models capable of understanding nuanced environmental and traffic scenarios with minimal human intervention.

Vision-Language Models (VLMs) have recently emerged as a compelling solution to this challenge. By aligning visual and textual modalities, VLMs demonstrate strong zero-shot reasoning capabilities across diverse domains without requiring extensive task-specific retraining \cite{Zhang2024, Li2025, Xu2024}. Built upon web-scale datasets, these models capture deep cross-modal correlations, enabling superior generalization and reduced reliance on manual labeling \cite{Zhang2024, Li2025, Xu2024}. State-of-the-art Large VLMs (LVLMs) have already outperformed conventional visual recognition models on tasks such as object detection, image classification, and semantic segmentation, proving effective not only in general computer vision benchmarks but also across downstream applications in scientific and industrial domains \cite{Zhang2024, Li2025, Xu2024, Yang2024KD}. Moreover, recent innovations extend VLM capabilities to handle dynamic, time-varying data, enabling models to reason over video sequences and capture evolving scenes and behaviors. These capabilities are essential for real-world deployment \cite{Kuo2024}.

Within the transportation domain, the application of VLMs has shown tremendous promise for semantic understanding of traffic environments. These models are increasingly being utilized to interpret complex scenes involving environmental context (e.g., fog, rain, snow), road surface conditions (e.g., dry, wet, flooded), vehicle behaviors, pedestrian interactions, and urban infrastructure \cite{Cao2024, Rivera2025, Zhang2024, Wang2023}. For example, integrating scene graphs into VLM pipelines significantly improves the model's ability to reason about causal relationships and event dynamics in traffic incidents, enabling deeper insights into accident events \cite{Lohner2024}. By leveraging multimodal inputs such as RGB imagery, thermal imaging, LiDAR point clouds, and high-definition map data, VLMs have demonstrated improved performance in understanding road scenes with greater accuracy and robustness \cite{Ashqar2024, Cao2024, Rivera2025, Zhang2024}. Crucially, the ubiquity of traffic surveillance camera networks across metropolitan and highway systems provides an abundant source of real-time data, yet these streams remain under-exploited. Integrating VLMs with existing video infrastructure offers unprecedented opportunities for real-time scene understanding, anomaly detection, and traffic incident summarization and reporting at scale \cite{Rivera2025, Kuo2024, Zhang2024}.

To further elevate the interpretive power of VLM-based systems, researchers have proposed multi-agent frameworks, wherein specialized agents collaboratively process different data modalities or subtasks \cite{Shriram2024, Xu2024, Kugo2025}. These architectures foster robustness, compositional reasoning, and parallel perception, which are particularly beneficial in dense or ambiguous traffic scenarios \cite{Shriram2024, Xu2024}. Recent work in Multi-Agent VQA demonstrates how foundation models distributed across agent roles can collectively outperform single-agent systems in zero-shot reasoning and question answering \cite{Jiang2024}. By decentralizing decision-making across cooperative agents, multi-agent VLMs improve generalization and robustness, which are critical for high-stakes applications such as autonomous driving, emergency response, and infrastructure safety monitoring \cite{Xu2024, Kugo2025}. These systems have also demonstrated proficiency in zero-shot detection of rare or hazardous events, offering scalable solutions for safety-critical deployments \cite{Shriram2024, Wang2023}.

Another essential component in VLM-powered transportation analysis is detailed and temporally aligned video captioning. High-quality captions enable the transformation of continuous visual input into structured, interpretable semantic representations, forming the basis for downstream tasks such as risk detection, abnormal behavior analysis, and driver assistance \cite{Kuo2024, Li2025, Bhooshan2022}. Techniques such as differential captioning and multimodal prompting with models like GPT-4V allow for granular descriptions that reflect intra-frame changes, temporal dependencies, and contextual interactions within traffic videos \cite{Kuo2024, Li2025}. These capabilities are especially important for automated monitoring systems that must parse fleeting events (e.g., abrupt lane changes, sudden stops, or near-collisions) with both accuracy and timeliness.

Despite their exceptional performance, large-scale VLMs pose significant computational demands, making them impractical for many real-world deployments, especially in embedded or resource-constrained systems. To bridge this gap, knowledge distillation (KD) has emerged as a vital strategy. KD techniques aim to transfer the reasoning and generalization power of a large, complex "teacher" model to a smaller, more efficient "student" model \cite{Xu2024, Yang2024KD}. While knowledge distillation focuses on transferring abstract reasoning abilities, skill distillation emphasizes replicating specific task-oriented behaviors,  both are highly relevant to transportation applications where timely and accurate decisions are paramount \cite{Xu2024, Yang2024KD}. Recent surveys underscore the role of KD in enabling lightweight VLMs that maintain high interpretability and performance, even under stringent latency or energy constraints \cite{Yang2024KD}.

Fine-tuned and distilled VLMs are increasingly being deployed in real-time traffic risk analysis pipelines. By generating automatic, semantically rich descriptions of road events, these models support rapid response, proactive management, and long-term planning by transportation authorities \cite{Wang2023, Rivera2025, Kuo2024}. Furthermore, automated risk reporting systems powered by VLMs can serve as intelligent roadside agents, continuously assessing road safety conditions and issuing timely alerts to prevent accidents. Such systems not only represent a significant leap toward Vision Zero goals but also enhance equity by enabling advanced traffic monitoring in resource-limited regions.

In this study, we present a novel multi-agent prompting and distillation framework designed to automatically generate high-fidelity traffic scene annotations and contextual risk assessments. This is achieved by orchestrating two popular VLMs (GPT-4o and o3-mini) using a structured Chain-of-Thought (CoT) strategy. The multi-perspective outputs from these expert agents serve as a knowledge-enriched supervision signals for supervised fine-tuning of a compact student VLM. Through this framework, we introduce VISTA (Vision for Intelligent Scene and Traffic Analysis ), a lightweight 3B model specifically engineered for understanding low-resolution traffic videos from existing traffic cameras and generating semantically faithful, risk-aware captions. Despite its substantially reduced parameter count, VISTA demonstrates strong competitive performance across commonly used metrics: BLEU-4, METEOR, ROUGE-L, and CIDEr, when quantitatively benchmarked against its larger teacher models. Importantly, its lightweight architecture enables efficient deployment on edge devices, thereby facilitating proactive risk monitoring without necessitating costly infrastructure upgrades. Our full training pipeline and model checkpoints can be found at \cite{Yang2025VISTA},.

\section{Data Description}
\begin{figure}[!ht]
    \centering
    \includegraphics[width=0.8\textwidth]{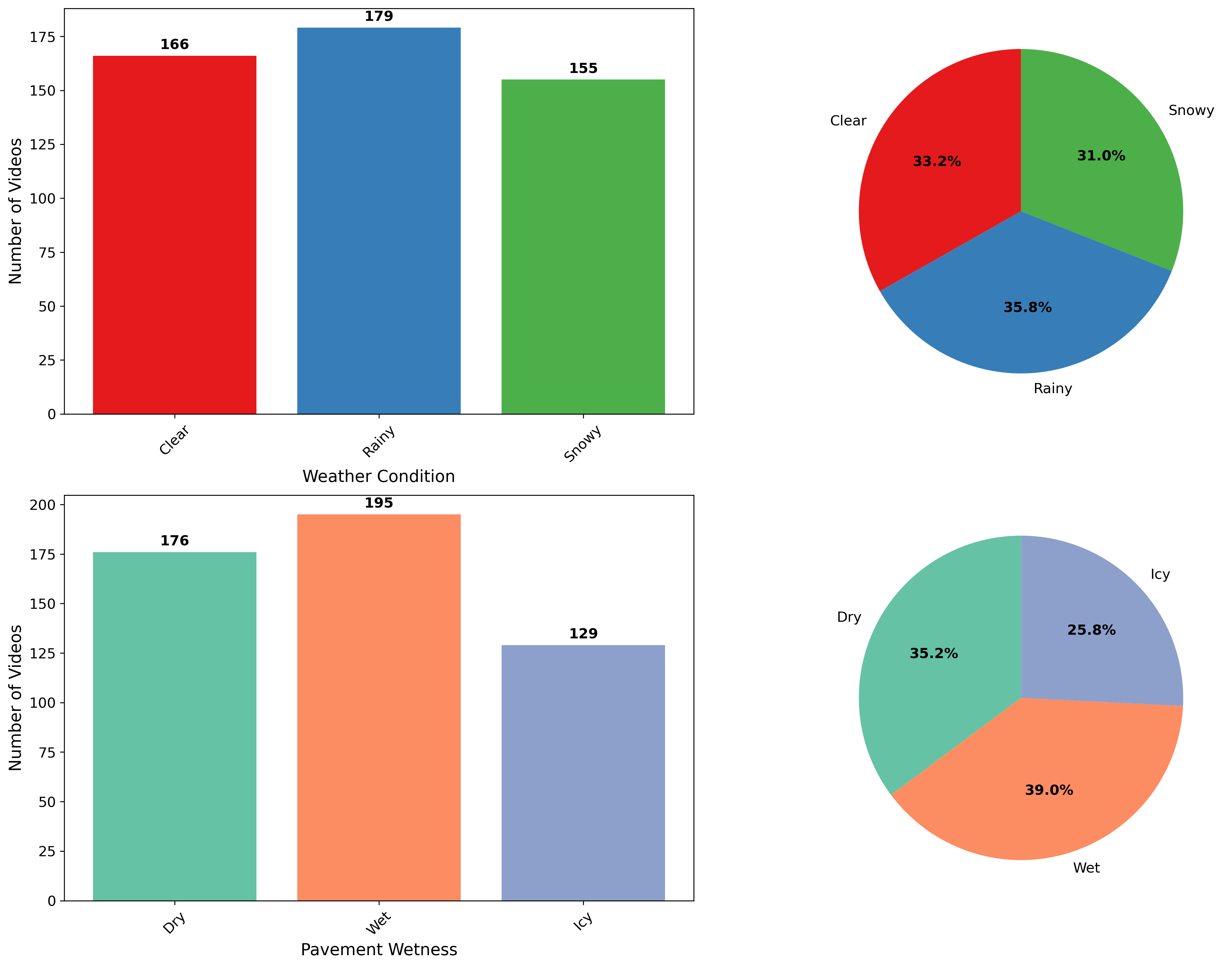}
    \caption{Distribution of the sampled training dataset.}
    \label{fig:data_count}
\end{figure}

We curated a large-scale, multi-modal dataset comprising synchronized video streams and road weather sensor data sourced from publicly accessible traffic cameras. The data were primarily collected from traffic monitoring systems in multiple states (e.g., Virginia, Georgia, and California) to ensure geographic diversity and a broad spectrum of environmental conditions. The collection process spanned from February 2025 to July 2025, capturing real-world traffic scenes across varying lighting, weather, and congestion scenarios. In total, over 21,000 short video clips were acquired, each ranging from 3 to 7 seconds in duration, depending on the source camera’s configuration. For model fine-tuning and controlled experimentation, we sampled 500 representative clips and analyzed their environmental and traffic-related distributions, as visualized in Figure~\ref{fig:data_count}.

\begin{figure}[!ht]
    \centering
    \includegraphics[width=1\textwidth]{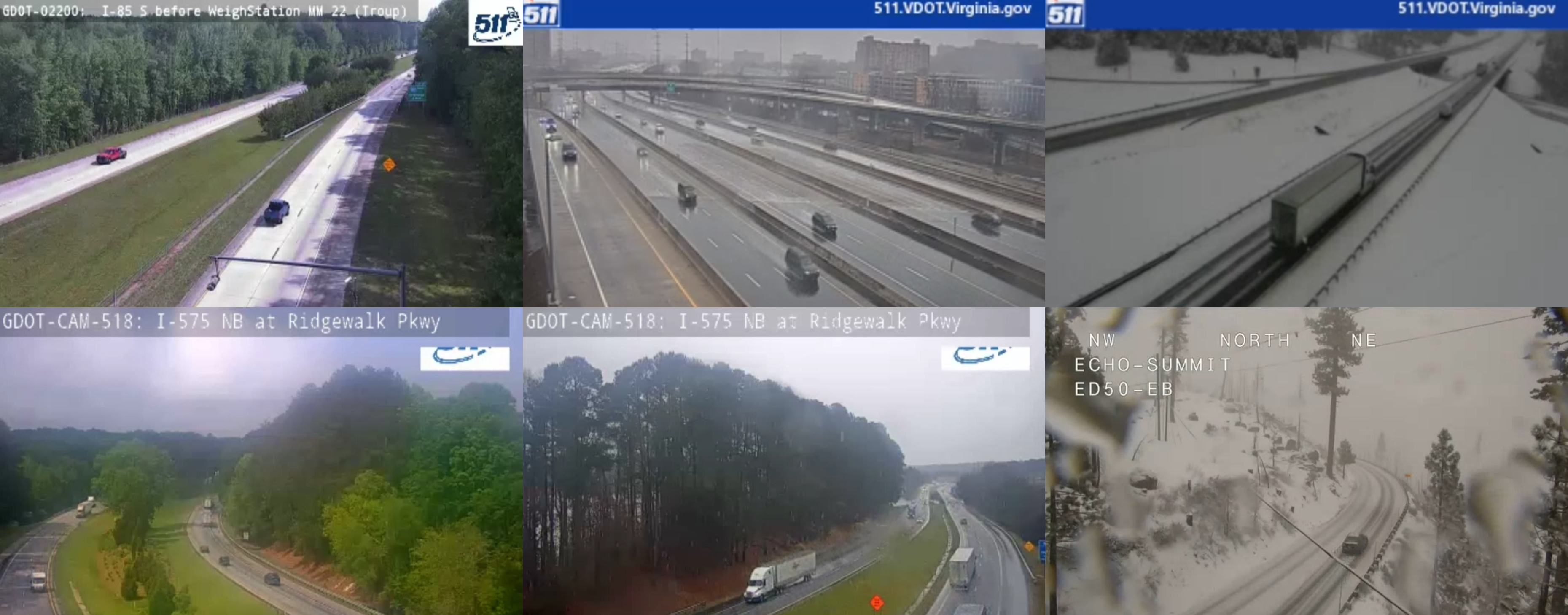}
    \caption{Sample frames for different weather/pavement wetness conditions: leftmost column – 1st row: clear with no precipitation, 2nd row: dry pavement; middle column – 1st row: rainy, 2nd row: wet pavement; rightmost column – 1st row: snowy, 2nd row: icy pavement.}
    \label{fig:data_frames}
\end{figure}

Figure~\ref{fig:data_frames} shows sample frames from the training dataset, illustrating a range of weather and pavement surface conditions. Notably, the majority of traffic cameras used for traffic monitoring are in relatively low resolutions. To evaluate model performance under realistic deployment conditions, we curated a test set comprising 200 video clips.

\section{Methodology}
\subsection{Overview}
To capture temporal dynamics while preserving computational efficiency, we extract multiple frames from each short traffic video clip at fixed intervals (typically 0.5 seconds). These frames are first processed by the GPT-4o model \cite{OpenAI2023}, which serves as Agent 1 in our multi-agent framework. Guided by a tailored CoT prompt \cite{Wei2022}, the model performs structured scene interpretation across six semantic dimensions: time of day (daytime or nighttime), road weather condition (clear, rainy, foggy, snowy), pavement surface wetness (dry, wet, flooded, snowy), vehicle behavior, traffic flow and speed, and congestion level. The output is a step-by-step, semantically rich caption that provides a comprehensive summary of the visual input.

Subsequently, both the selected video frames and the structured scene analysis from Agent 1 are passed to the o3-mini model \cite{OpenAI2025}, which functions as Agent 2 for reasoning. This agent employs a dedicated CoT prompt to perform traffic risk interpretation across multiple safety-critical dimensions, including environmental risk factors, vehicle behavior risk, traffic flow risk, and overall safety level, alongside actionable insights such as driver alerts and recommended safe speeds.

\begin{figure}[!ht]
    \centering
    \includegraphics[width=1\textwidth]{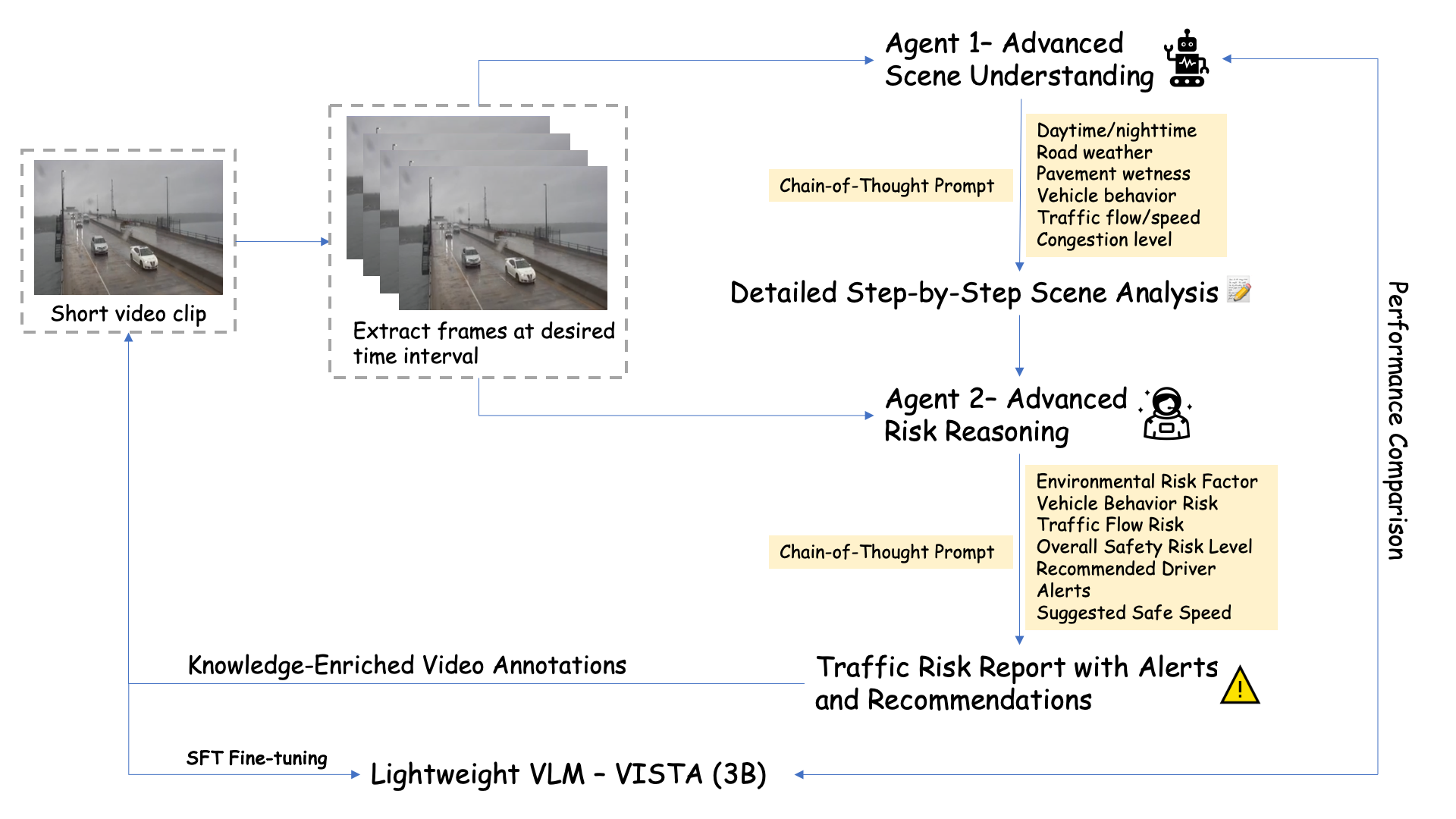}
    \caption{Overview of the proposed framework.}
    \label{fig:framework}
\end{figure}

The combined output from both agents, consisting of detailed scene understanding and structured risk assessment, serves as high-quality, knowledge-enriched pseudo-annotations. These annotations are then used to perform supervised fine-tuning (SFT) on a compact 3B Qwen2.5-VL model \cite{Qwen2.5-VL}, leading to our distilled student model, VISTA. which can be deployed for real-time traffic safety applications. An overview of the training and inference pipeline is illustrated in Figure~\ref{fig:framework}.

\subsection{CoT Prompt Design for Multi-Agent Video Analysis}

To elicit structured and context-rich outputs from VLMs, we design two customized CoT prompts tailored to each expert agent’s specific capabilities. These prompts guide the reasoning steps for interpreting visual scenes and assessing traffic risks, resulting in coherent, interpretable outputs usable for downstream supervision.

\subsubsection*{Superior Road Scene Understanding}

\begin{figure}[!ht]
    \centering
    \includegraphics[width=\linewidth]{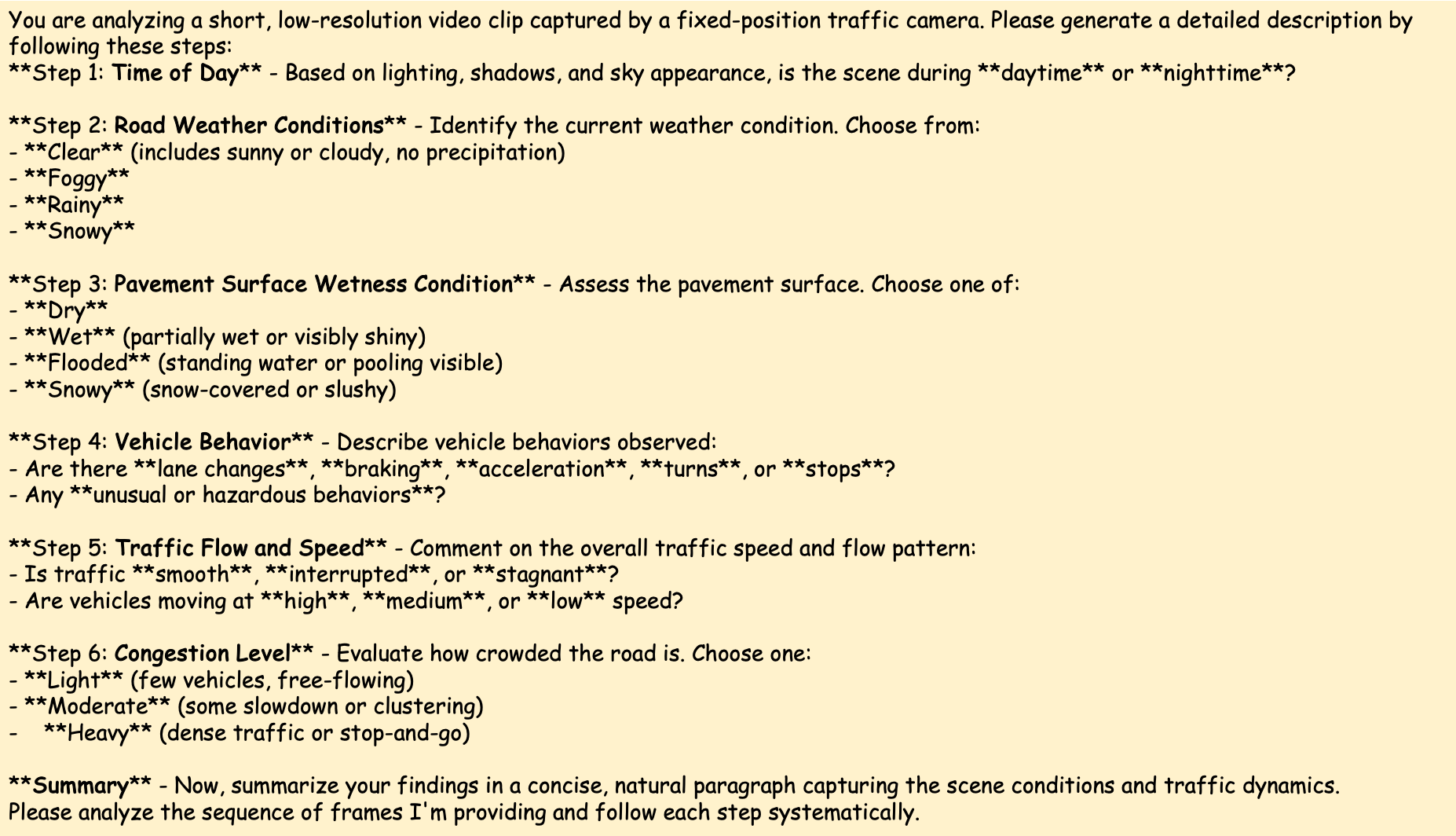}
    \caption{CoT prompt design for Agent 1 (GPT-4o): structured scene understanding.}
    \label{fig:cot_agent1}
\end{figure}

Agent 1 (GPT-4o) is responsible for extracting high-level semantic information from short video clips by following a multi-step CoT prompt (see Figure~\ref{fig:cot_agent1}). This prompt breaks down the video interpretation into six semantic dimensions:

\begin{enumerate}
    \item \textbf{Time of Day}: Distinguishing between daytime and nighttime based on ambient light and sky conditions. 
    \item \textbf{Road Weather Conditions}: Classifying the environment as clear, foggy, rainy, or snowy.
    \item \textbf{Pavement Wetness Condition}: Assessing the road surface as dry, wet (shiny or moist), flooded (pooled water), or snowy (slush/coverage). 
    \item \textbf{Vehicle Behavior}: Identifying maneuvers such as lane changes, braking, acceleration, turns, or unusual/hazardous actions. 
    \item \textbf{Traffic Flow and Speed}: Estimating traffic smoothness and the general speed level (high, medium, or low). 
    \item \textbf{Congestion Level}: Categorizing congestion as light, moderate, or heavy.
\end{enumerate}

\begin{figure}[!ht]
    \centering
    \includegraphics[width=\linewidth]{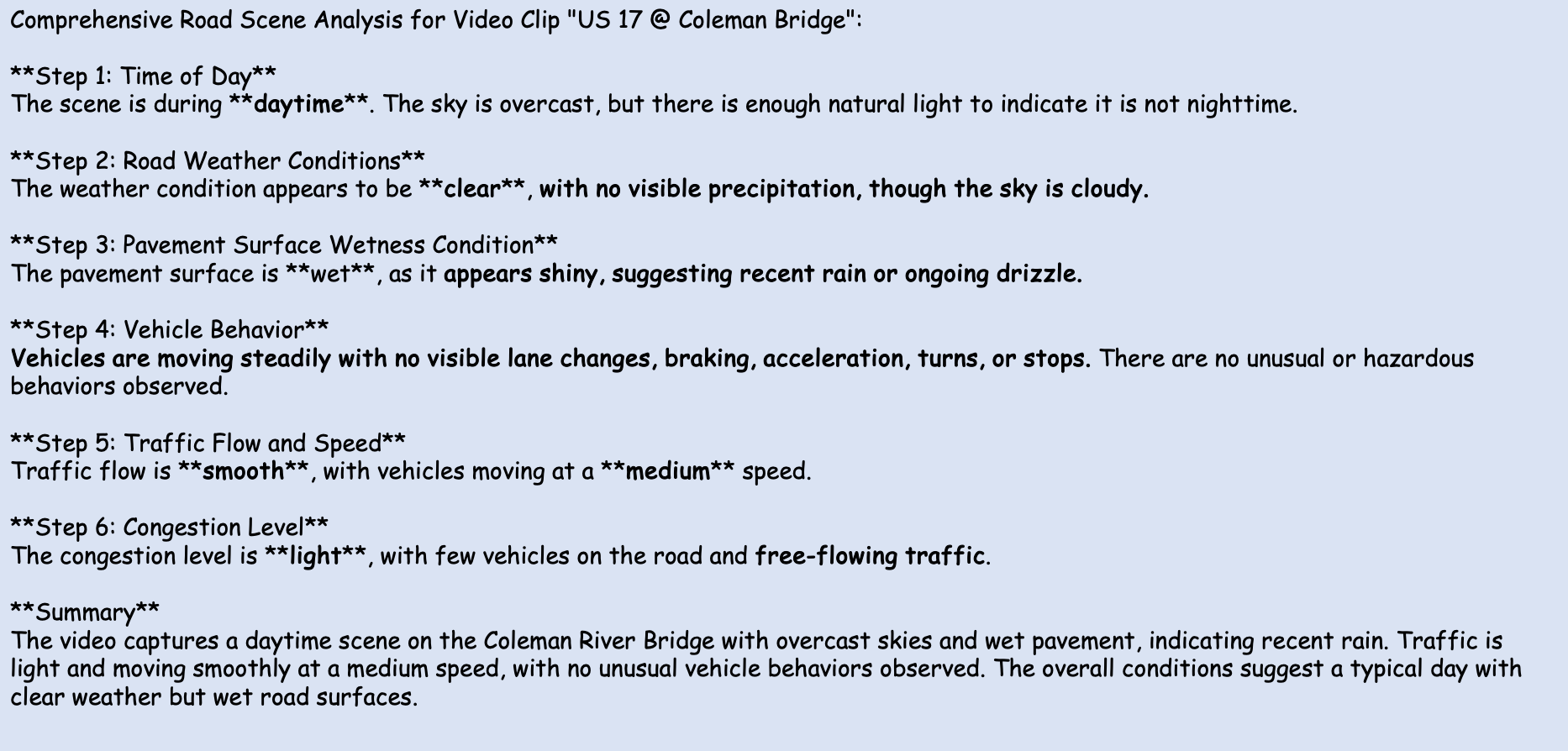}
    \caption{Sample output from Agent 1 (GPT-4o): multi-step scene analysis based on visual cues and traffic dynamics.}
    \label{fig:agent1_output}
\end{figure}

By decomposing the analysis in this manner, the prompt encourages the VLM to produce a logically ordered and granular description. The output is not a generic caption but a semantically annotated scene summary, concluding with a synthesized paragraph summarizing the overall traffic conditions. A sample output is shown in Figure~\ref{fig:agent1_output}.

\subsubsection*{Advanced Risk Reasoning}

\begin{figure}[!ht]
    \centering
    \includegraphics[width=\linewidth]{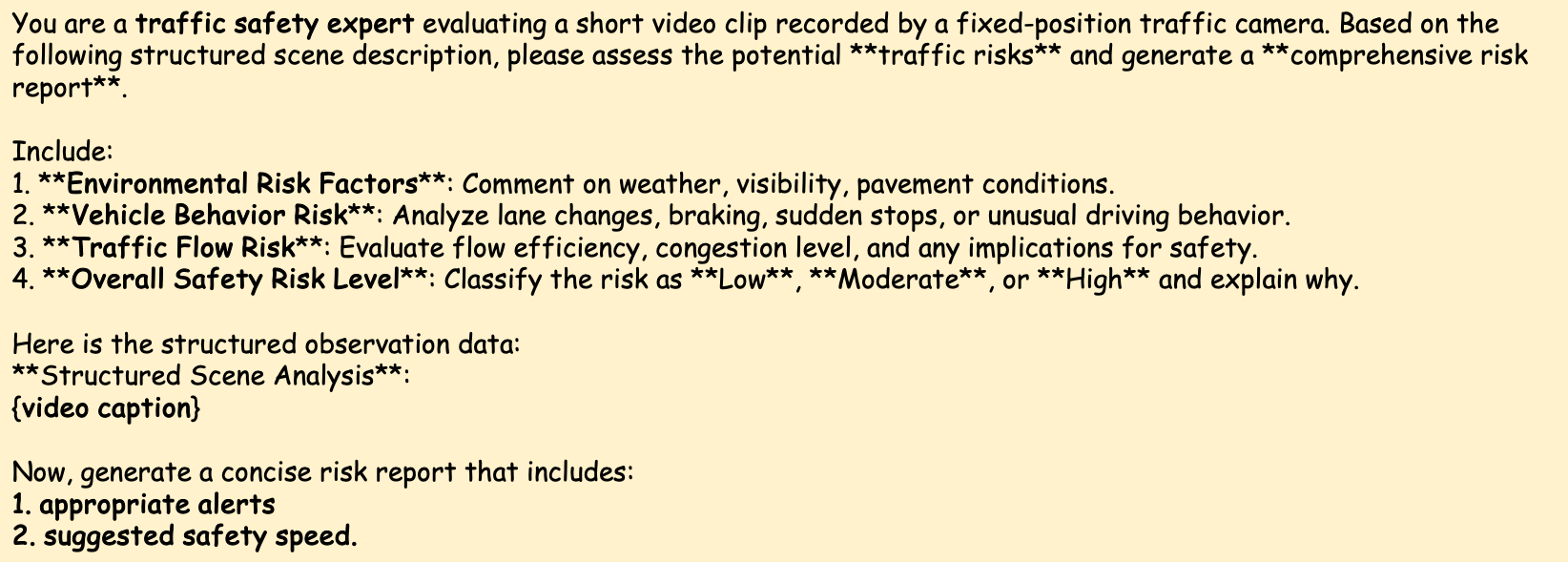}
    \caption{CoT prompt design for Agent 2 (o3-mini): structured traffic risk reasoning.}
    \label{fig:cot_agent2}
\end{figure}

Building upon the structured scene analysis from Agent 1, the second agent (o3-mini) performs risk interpretation using a dedicated CoT prompt (see Figure~\ref{fig:cot_agent2}). The prompt positions the model as a traffic safety expert and requests a comprehensive report in four dimensions:

\begin{enumerate}
    \item \textbf{Environmental Risk Factors}: This component focuses on analyzing the interplay between visibility, weather, and pavement surface conditions—three critical elements in assessing traffic safety. Time of day plays a decisive role in determining visibility, with nighttime or low-light conditions significantly impairing a driver's ability to detect road hazards, pedestrians, or surface anomalies such as standing water or debris. Weather factors such as rain, fog, or snow further compound these risks by introducing glare, reduced contrast, or obscured features. Pavement wetness, in particular, poses substantial safety concerns by affecting vehicle traction, braking distance, and hydroplaning likelihood. For example, a reflective road surface under overcast skies may suggest recent precipitation, while visible pooling indicates potential flooding. Distinguishing between partially wet and deeply saturated pavement is therefore crucial for anticipating vehicle instability and enabling downstream risk prediction. The CoT prompt guides the model to reason across these dimensions collectively, enabling a comprehensive assessment of how environmental conditions influence roadway safety. 
    \item \textbf{Vehicle Behavior Risk}: This dimension evaluates whether the observed driving patterns suggest cautious or erratic behavior, offering insight into potential latent hazards. Behavioral responses such as sudden braking, abrupt acceleration, or frequent lane changes often reflect driver reactions to perceived risks—such as reduced visibility, surface irregularities, or obstructions not directly captured in the visual field. Importantly, clusters of such evasive maneuvers across multiple vehicles may signal the presence of high-risk zones ahead, including flooded pavement, debris fields, or stalled traffic. By analyzing both the frequency and distribution of these maneuvers, the model can infer emergent risk factors and support preemptive safety reasoning that extends beyond the immediate visual context.
    \item \textbf{Traffic Flow Risk}: This component assesses the stability and efficiency of traffic flow to identify dynamic risk patterns. Consistent, smooth flow typically indicates low interaction risk, whereas abrupt fluctuations in vehicle speed or spacing may signal unexpected environmental disturbances—such as roadway obstructions, water accumulation, or sudden visibility drops. Such disruptions not only degrade flow efficiency but also elevate the probability of rear-end collisions, particularly under conditions of limited traction or poor visual perception. The CoT prompt enables the model to reason temporally, detecting irregularities in flow continuity and interpreting them as early indicators of potential hazards ahead. This temporal perspective is critical for proactive traffic risk evaluation.
    \item \textbf{Overall Safety Risk Level}: Providing a low/moderate/high risk classification with justification.
\end{enumerate}

Additionally, the prompt requests actionable driver guidance in the form of:
\begin{itemize}
    \item \textbf{Alerts}: Warnings or advisories relevant to current road conditions.
    \item \textbf{Suggested Safety Speed}: A recommended driving speed that reflects current visibility, pavement, and flow characteristics.
\end{itemize}

\begin{figure}[!ht]
    \centering
    \includegraphics[width=\linewidth]{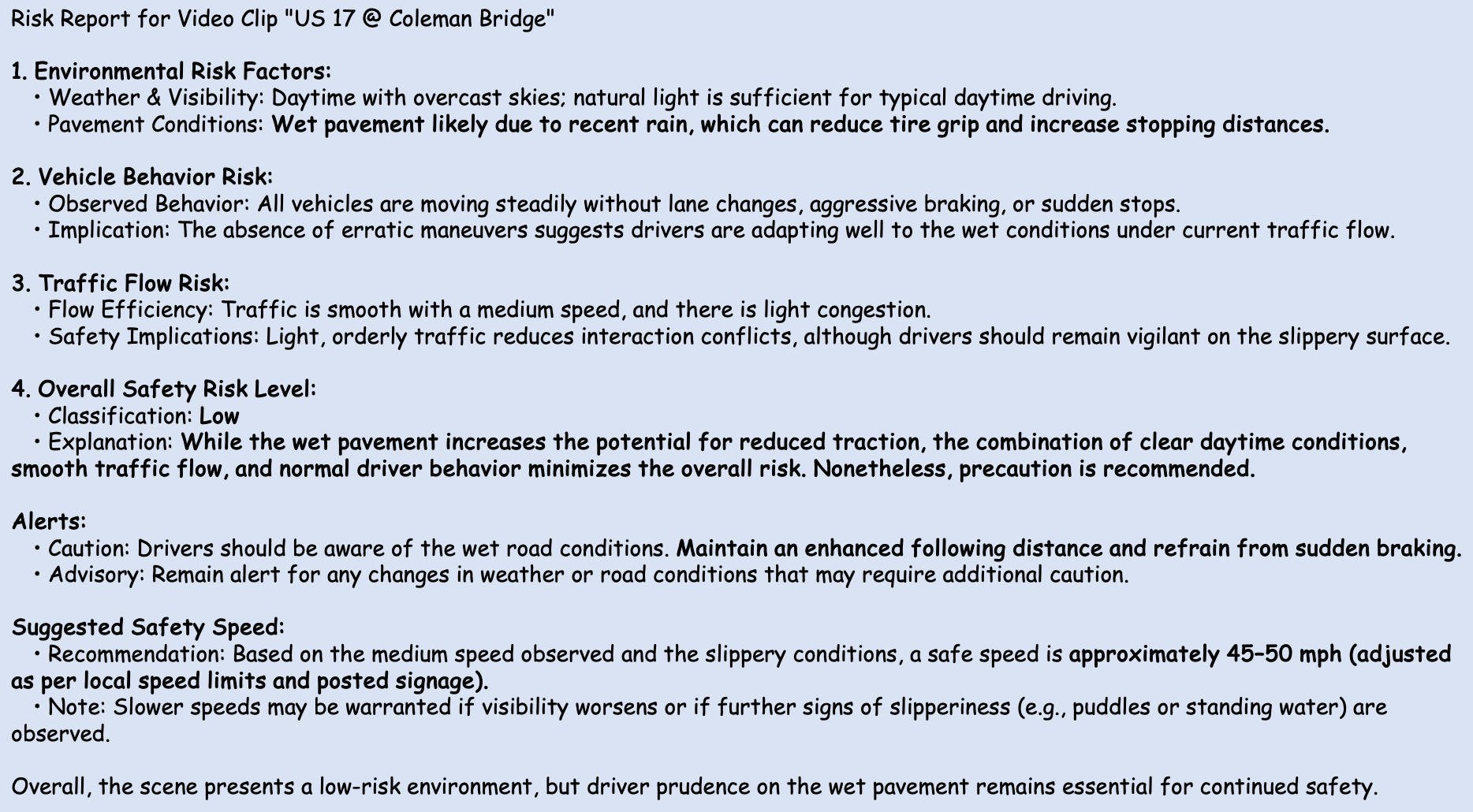}
    \caption{Sample output from Agent 2 (o3-mini): structured risk analysis and advisory reasoning based on scene context.}
    \label{fig:agent2_output}
\end{figure}

This structured reasoning output represents an interpretable risk abstraction over raw video frames. A representative sample response generated by o3-mini is visualized in Figure~\ref{fig:agent2_output}.

\subsubsection*{Prompt Structure for General-Purpose Fine-Tuning}

While the previous sections describe the two structured prompts used during knowledge-enriched label generation, our general-purpose training requires a unified prompt format to supervise the student model directly. This ensures that the model learns not only how to reason but also structure its outputs in an interpretable, standardized format compatible with downstream risk assessment tasks.

\begin{figure}[!ht]
    \centering
    \includegraphics[width=\linewidth]{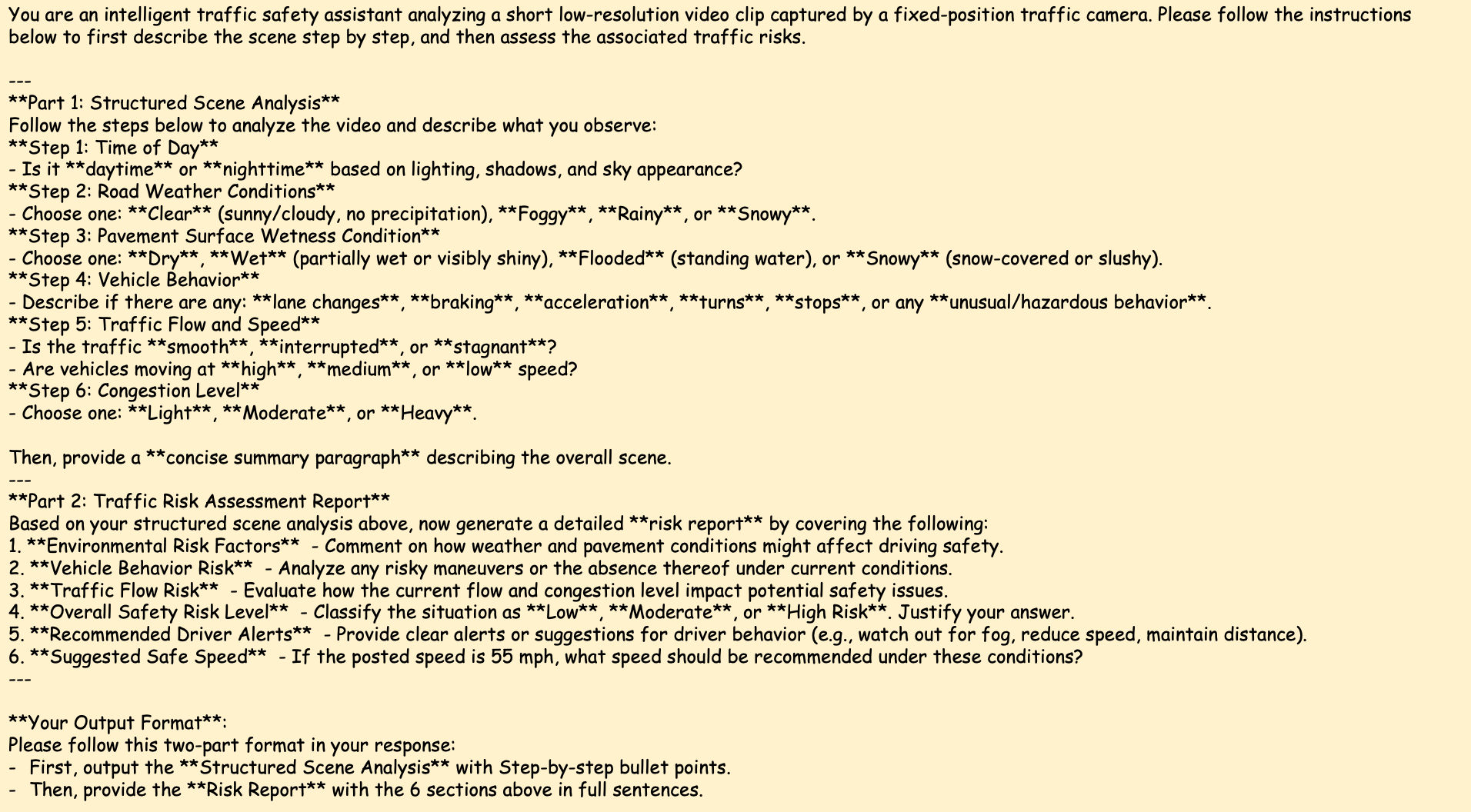}
    \caption{Unified training-time prompt used to supervise the student model during supervised fine-tuning.}
    \label{fig:training_prompt}
\end{figure}

Figure~\ref{fig:training_prompt} presents the full training-time instruction used for the training of VISTA. The prompt integrates the elements from the structured prompts of the two teacher agents and explicitly positions the student model as an intelligent traffic safety assistant tasked with generating a two-part response: (1) a structured scene analysis, and (2) a traffic risk report. This format blends descriptive grounding (scene perception) with high-level reasoning (risk abstraction), thereby encouraging the model to encode both factual observations and safety-critical interpretations.

\subsection{Multi-Agent Knowledge Distillation via Caption and Risk Label Generation}

To enable a lightweight VLM to perform high-quality road scene interpretation and traffic risk inference, we propose a novel \textit{multi-agent knowledge distillation} framework. Unlike conventional approaches that distill feature representations or logits from a single teacher model, our method orchestrates multiple expert agents, i.e., GPT-4o and o3-mini, to generate rich, semantically-structured supervision signals in the form of captions and risk reports. These outputs are subsequently used to supervise a smaller VLM via supervised fine-tuning (SFT).

\subsubsection*{Distillation Setup}

Let $\mathrm{VideoSet} = \{v_i\}_{i=1}^N$ denote a collection of $N$ short traffic video clips. For each video $v_i$, we extract a sequence of frames denoted by $\mathrm{Frames}_i = \{f_i^{(1)}, f_i^{(2)}, \ldots, f_i^{(T)}\}$ at uniform temporal intervals (e.g., 0.5 seconds). 

The extracted $\mathrm{Frames}_i$ are first processed by Agent~1 (GPT-4o) with a CoT prompt, resulting in a structured scene description:
\begin{equation}
    \hat{y}_i^{\mathrm{scene}} = \mathrm{GPT\text{-}4o}(\mathrm{Frames}_i;\, \mathrm{Prompt}_1),
\end{equation}
where $\mathrm{Prompt}_1$ is designed to infer attributes such as daytime/nighttime, weather condition, pavement wetness, traffic flow, vehicle behavior, and congestion level.

The scene caption $\hat{y}_i^{\mathrm{scene}}$ is then combined with the original frames and passed to Agent~2 (o3-mini) for advanced risk reasoning:
\begin{equation}
    \hat{y}_i^{\mathrm{risk}} = \mathrm{o3\text{-}mini}([\mathrm{Frames}_i,\, \hat{y}_i^{\mathrm{scene}}];\, \mathrm{Prompt}_2),
\end{equation}
where $\mathrm{Prompt}_2$ targets traffic risk interpretation, including environmental risk, behavior risk, traffic flow-related risk, an overall safety assessment, and actionable driving recommendations (alerts and safe speed).

\subsubsection*{Unified Label Construction}

For each video $v_i$, the outputs from Agent~1 (scene-level description) and Agent~2 (risk-level reasoning) are concatenated at the text level to form a unified annotation:
\begin{equation}
    \tilde{y}_i = \mathrm{concat}(\hat{y}_i^{\mathrm{scene}},\; \hat{y}_i^{\mathrm{risk}}),
\end{equation}
where $\hat{y}_i^{\mathrm{scene}}$ and $\hat{y}_i^{\mathrm{risk}}$ are the raw natural language outputs from the two expert agents.  
This unified text sequence $\tilde{y}_i$ encapsulates both perceptual understanding and prescriptive reasoning about the traffic scene.

Before training, $\tilde{y}_i$ is tokenized into a sequence of discrete token IDs:
\begin{equation}
    y_i = \mathrm{Tokenizer}(\tilde{y}_i) = \left[y_i^{(1)}, y_i^{(2)}, \dots, y_i^{(L_i)}\right],
\end{equation}
where $L_i$ is the tokenized sequence length.

\subsubsection*{Distillation Objective}

Let the student model be a lightweight VLM, i.e., our VISTA model, denoted as $\mathrm{VISTA}_\theta$, where $\theta$ are the model parameters. In practice, both teacher agents output discrete text rather than token-level probability distributions. Thus, our distillation reduces to \textit{supervised fine-tuning} on tokenized pseudo-labels, where the student is trained to maximize the likelihood of reproducing the teacher-generated annotations.

Formally, given the input frame sequence $\mathrm{Frames}_i$, the model generates a sequence of logits:
\begin{equation}
    \mathbf{z}_i^{(t)} = \mathrm{VISTA}_\theta(\mathrm{Frames}_i, y_i^{(<t)}), \quad t = 1, \dots, L_i,
\end{equation}
where $y_i^{(<t)}$ denotes the tokens preceding position $t$.  
The logits $\mathbf{z}_i^{(t)} \in \mathbb{R}^V$ (with $V$ being the vocabulary size) are transformed into probability distributions via the softmax:
\begin{equation}
    p_\theta^{(t)} = \mathrm{softmax}(\mathbf{z}_i^{(t)}).
\end{equation}

The sequence-level cross-entropy loss for SFT is:
\begin{equation}
    \mathcal{L}_{\mathrm{CE}} = - \frac{1}{L_i} \sum_{t=1}^{L_i} \log p_\theta^{(t)}[y_i^{(t)}],
\end{equation}
where $p_\theta^{(t)}[y_i^{(t)}]$ denotes the predicted probability assigned to the ground-truth token at step $t$.

The overall fine-tuning objective over the dataset $\mathcal{D}$ is:
\begin{equation}
    \mathcal{L}_{\mathrm{SFT}} = \frac{1}{|\mathcal{D}|} \sum_{i=1}^{|\mathcal{D}|} \mathcal{L}_{\mathrm{CE}}\left(\mathrm{VISTA}_\theta(\mathrm{Frames}_i),\, y_i\right).
\end{equation}

This knowledge distillation strategy enables the transfer of rich, structured knowledge from large-scale VLMs to a compact student model. GPT-4o contributes detailed visual understanding, while o3-mini provides contextual and risk-aware reasoning. Their distinct, yet complementary expertise allows the student model to learn and generalize beyond descriptive captioning to safety-critical reasoning and interpretation.

\subsection{Training and Evaluation Pipeline}

\begin{figure}[!ht]
    \centering
    \includegraphics[width=1\linewidth]{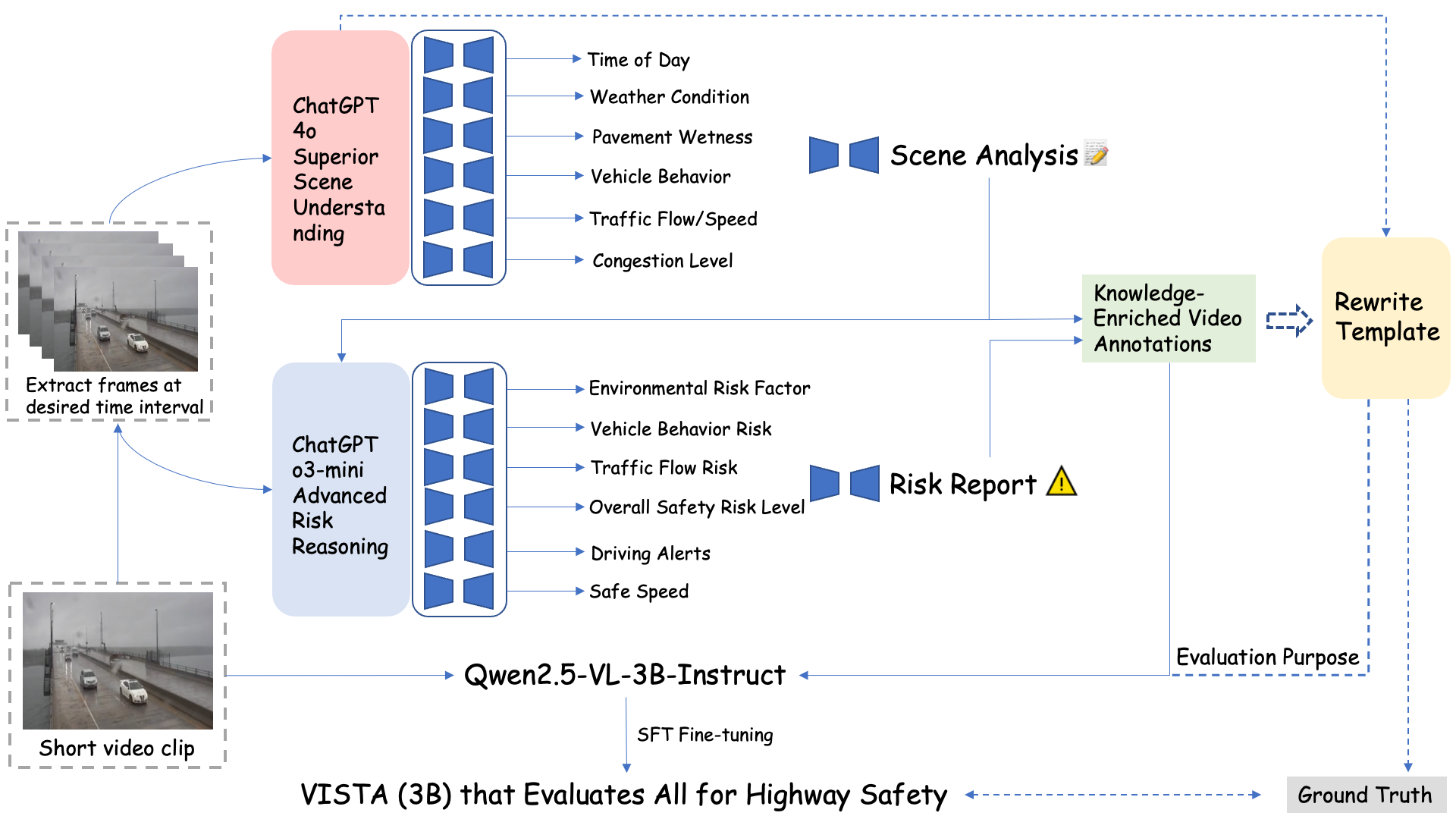}
    \caption{Overall pipeline of multi-agent supervision, fine-tuning, and evaluation for VISTA.}
    \label{fig:pipeline}
\end{figure}

Figure~\ref{fig:pipeline} illustrates the end-to-end training and evaluation framework of VISTA. To fully leverage the strengths of LVLMs while enabling lightweight deployment, our pipeline consists of two supervision paths:

\textbf{(1) Multi-Agent Knowledge-Enriched Supervision.} The first path generates rich, semantically grounded annotations from a pair of expert LVLMs: GPT-4o and o3-mini. Specifically, We sample frames from short video clips and pass them along with two specialized prompting branches: (i) GPT-4o for fine-grained scene analysis and captioning, and (ii) o3-mini for contextual risk reasoning. Their outputs are merged and organized into structured, knowledge-enriched video annotations which are used as pseudo-labels to supervise the fine-tuning of a 3B student model (Qwen2.5-VL-3B-Instruct), leading to our VISTA model, suitable for deployment in real-world highway safety monitoring.

\begin{figure}[!ht]
    \centering
    \includegraphics[width=0.6\linewidth]{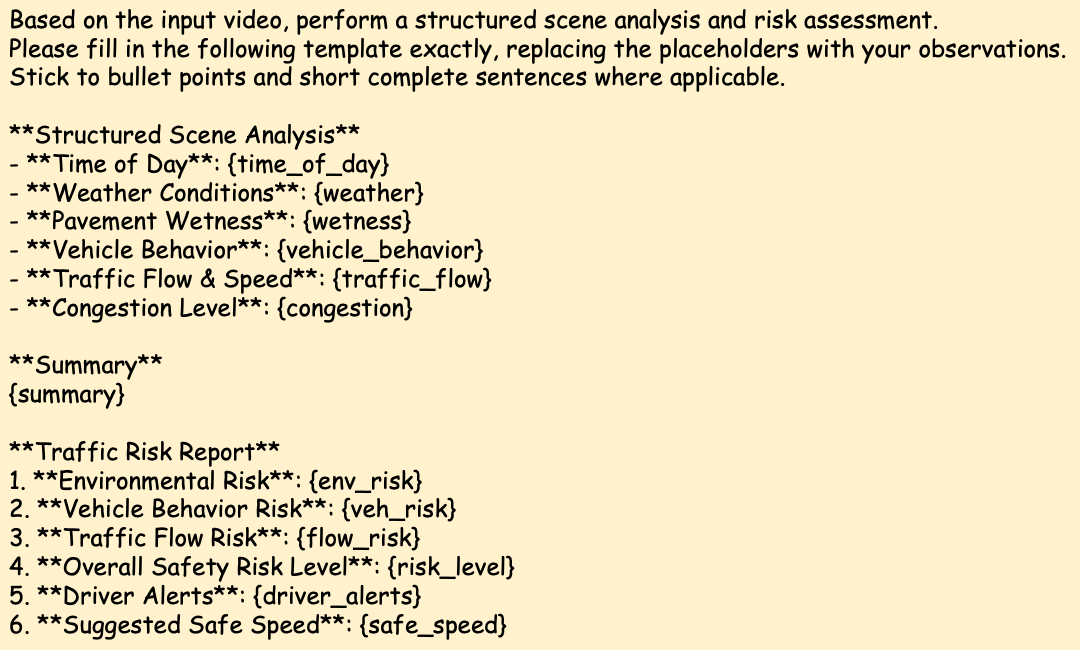}
    \caption{Rewritten template for unified evaluation.}
    \label{fig:template}
\end{figure}

\textbf{(2) Template-Based Evaluation.} To rigorously assess the semantic fidelity of VISTA against its LVLM teachers, we construct another parallel supervision path. The multi-agent outputs are reformatted into a unified evaluation template (refer to Figure \ref{fig:template}) that reflects the exact linguistic structure expected during inference. We then further fine-tune the same 3B model under this template supervision using identical hyperparameters. This guarantees structural alignment between model outputs and GPT-4o reference generations, enabling precise metric-based comparison.

\textbf{Testing Protocol.} At inference time, we apply the same rewritten template to all test video clips using GPT-4o to produce ground-truth reference outputs. In parallel, our fine-tuned VISTA is evaluated on the same video clips. Performance is assessed over 200 samples using BLEU-4, METEOR, ROUGE-L, and CIDEr, as detailed in Section Evaluation Metrics. The resulting scores (Table~\ref{tab:ablations}) quantify the alignment between the distilled lightweight model and its large-scale teacher ensemble.

\section{Experiments and Results}

In this section, we evaluate the performance of our lightweight VLM model, VISTA, under different fine-tuning schemes taht target distinct model components, with all fine-tuned variants compared against the pretrained model as a baseline. All model variants are evaluated against GPT-4o outputs as reference on a held-out test set comprising 200 traffic video clips. These videos are curated to be representative of diverse real-world scenarios involving different weather conditions, pavement wetness levels, traffic flow conditions and dynamics.

\subsection{Evaluation Metrics}

Our goal is to assess both local n-gram fidelity and global semantic alignment. For this purpose, we employ four widely adopted evaluation metrics from the captioning literature: BLEU-4, METEOR, ROUGE-L, and CIDEr. Each metric captures a different facet of linguistic similarity, enabling a holistic comparison.

\vspace{0.5em}
\noindent\textbf{BLEU-4} \cite{Papineni2002} evaluates n-gram precision, focusing on exact phrase-level overlap between the predicted caption and the reference. Specifically, BLEU-4 computes the modified precision of 4-grams as:

\begin{equation}
\mathrm{BLEU\text{-}4} = \mathrm{BP} \cdot \exp\left( \sum_{n=1}^{4} w_n \cdot \log p_n \right),
\end{equation}

where $p_n$ denotes the modified precision for n-grams (up to 4), $w_n = \frac{1}{4}$, and $\mathrm{BP}$ is the brevity penalty to discourage overly short predictions. In our case, we compute BLEU-4 between each VISTA-generated caption $\hat{y}_i$ and the corresponding GPT-4o reference $y_i$ over all 200 samples and report the average.

\vspace{0.5em}
\noindent\textbf{METEOR} \cite{Banerjee2005} balances unigram precision and recall, incorporating stemming, synonymy, and word order through alignment. It is defined as:

\begin{equation}
\mathrm{METEOR} = F_{\mathrm{mean}} \cdot (1 - P),
\end{equation}

where $F_{\mathrm{mean}}$ is the harmonic mean of unigram precision and recall, and $P$ is a penalty based on the fragmentation of matched chunks. This metric tends to correlate better with human judgment, especially for short descriptive text such as video captions.

\vspace{0.5em}
\noindent\textbf{ROUGE-L} \cite{Lin2004} measures sentence-level structural similarity based on the 
Longest Common Subsequence (LCS) between the predicted and reference token sequences. 
Let $\hat{y}_i=\{\hat{y}_i^{(t)}\}_{t=1}^{|\hat{y}_i|}$ denote the prediction and 
$y_i=\{y_i^{(t)}\}_{t=1}^{|y_i|}$ the reference. Define
\[
\mathrm{LCS}_i = \mathrm{LCS}(\hat{y}_i, y_i),
\]
the length of their longest common subsequence. From this, the LCS-based recall and precision are:
\[
R^{\mathrm{LCS}}_i=\frac{\mathrm{LCS}_i}{|y_i|},\qquad
P^{\mathrm{LCS}}_i=\frac{\mathrm{LCS}_i}{|\hat{y}_i|}.
\]
Following the official \texttt{rouge\_score} implementation, we report the \emph{F1-based ROUGE-L score} 
($\beta=1$), which harmonically combines LCS-based precision and recall:
\[
\mathrm{ROUGE\text{-}L}_i
= \frac{2\,R^{\mathrm{LCS}}_i\,P^{\mathrm{LCS}}_i}
       {R^{\mathrm{LCS}}_i+\beta^2P^{\mathrm{LCS}}_i}.
\]
The final metric is obtained by averaging over all test clips.

\vspace{0.5em}
\noindent\textbf{CIDEr} \cite{Vedantam2015} evaluates consensus across multiple references by computing TF-IDF weighted n-gram similarity. For each caption, it is defined as:

\begin{equation}
\mathrm{CIDEr} = \frac{1}{N} \sum_{i=1}^{N} \sum_{n=1}^{4} w_n \cdot \mathrm{sim}_{\mathrm{TF\text{-}IDF}}^n(\hat{y}_i, y_i),
\end{equation}

where $w_n$ is the weight for n-gram $n$ (typically uniform), and $\mathrm{sim}_{\mathrm{TF\text{-}IDF}}^n$ is the cosine similarity between TF-IDF vectors of the n-grams. Since we use single-reference evaluation (GPT-4o output), CIDEr provides insight into descriptive richness and term relevance.

\vspace{0.5em}
\noindent\textbf{Composite Score.} To facilitate unified comparison, we compute a composite score from all four metrics, defined as:

\begin{equation}
\label{eq:score}
\mathrm{Score} = \frac{\mathrm{BLEU\text{-}4} + \mathrm{METEOR} + \mathrm{ROUGE\text{-}L} + 0.1 \times \mathrm{CIDEr}}{4} \times 100,
\end{equation}

where CIDEr is scaled by 0.1 to normalize its value range. This aggregate formulation captures both structural accuracy and semantic fidelity while accounting for redundancy and term frequency. A similar multi-metric strategy has been adopted in recent literature such as CityLLaVA \cite{Deng2024CityLLaVA}, where BLEU, METEOR, and ROUGE are combined to evaluate semantic grounding and coherence in urban scene descriptions. In our case, the composite score enables a holistic assessment of model performance, integrating both exact n-gram matches (BLEU), semantic fluency and recall (METEOR), sentence-level structure (ROUGE-L), and consensus with human-like outputs (CIDEr).

\subsection{Implementation Details}

We implement our full pipeline using the Qwen2.5-VL-3B architecture, a compact yet capable instruction-tuned vision-language model available through HuggingFace. All training is conducted on a high-performance server equipped with four NVIDIA A6000 GPUs (each with 48GB memory). We adopt mixed-precision training using bfloat16 for memory efficiency and enable DeepSpeed ZeRO Stage 3 parallelism and gradient checkpointing to scale training across longer input contexts without exceeding GPU memory constraints.

\textbf{Training Setup.} The training follows a consistent hyperparameter configuration across all model variants. We use a cosine learning rate scheduler with an initial learning rate of $2 \times 10^{-6}$ and a warm-up ratio of 0.03. The per-device batch size is set to 1, with gradient accumulation steps of 1. Inputs are truncated at a maximum sequence length of 8,192 tokens, while image resolution is fixed at $224 \times 224$, bounded by a total pixel limit of 50,176 to avoid memory overflow. Each training run is conducted for 5 epochs, and checkpoints are saved every 1,000 steps, with only the latest checkpoint retained.

\textbf{Component-Wise Fine-Tuning.} We perform end-to-end fine-tuning of all key components in the Qwen2.5-VL-3B architecture:
\begin{itemize}
    \item \textit{Visual Encoder:} The CLIP-style backbone is updated to specialize in low-resolution traffic footage under diverse environmental conditions.
    \item \textit{Language Decoder:} Tuning the LLM decoder enables the model to emulate expert-style reasoning and reporting formats derived from large teacher models (GPT-4o and o3-mini).
    \item \textit{Cross-Modal MLP Fusion:} This module aligns vision and language representations. Fine-tuning it improves grounding of visual semantics into structured CoT outputs.
\end{itemize}

While parameter-efficient fine-tuning (PEFT) strategies \cite{Ding2023} such as LoRA \cite{Hu2022} and adapter modules are effective in general-purpose scenarios, we deliberately opt for full-model fine-tuning. This decision is driven by the significant domain gap between pretraining corpora and our targeted deployment setting:rural traffic environments with complex road/weather dynamics and low-visibility conditions. In our empirical assessment, adapter-based methods struggled to adapt cross-modal dependencies and preserve fine-grained semantics under such domain shift. Full fine-tuning ensures holistic adaptation of the vision-language model, which is essential for capturing subtle visual cues (e.g., glare from wet pavement, faint lane markings) and delivering high-fidelity safety reasoning required in ITS applications.

\subsection{Results}

Table~\ref{tab:ablations} reports the performance of the original pretrained model and four fine-tuned model variants on the same test set of 200 traffic video clips. Evaluation is conducted against reference outputs produced by GPT-4o using four standard captioning metrics (BLEU-4, METEOR, ROUGE-L, and CIDEr), along with a composite score as defined in Section: Evaluation Metrics.

\begin{table}[!ht]
    \caption{Comparison of model performance under different fine-tuning schemes.}
    \label{tab:ablations}
    \begin{center}
        \begin{tabular}{l c c c c c}
            \hline
            \textbf{Model} & \textbf{BLEU-4} & \textbf{METEOR} & \textbf{ROUGE-L} & \textbf{CIDEr} & \textbf{Score} \\
            \hline
            \textit{3B original}                 & 0.2517 & 0.5396 & 0.3902 & 0.2984 & 30.28 \\
            \textit{3B mlp}                      & 0.2581 & 0.5287 & 0.4040 & 0.3363 & 30.61 \\
            \textit{3B mlp+vision}               & 0.2722 & 0.5281 & 0.4346 & 0.2413 & 31.48 \\
            \textit{3B mlp+llm}                  & 0.3269 & \textbf{0.5691} & 0.4862 & 0.6712 & 36.23 \\
            \textit{3B llm+mlp+vision} (VISTA)   & \textbf{0.3289} & 0.5634 & \textbf{0.4895} & \textbf{0.7014} & \textbf{36.30} \\
            \hline
        \end{tabular}
    \end{center}
    \vspace{1.5mm}
    {\footnotesize
    \textbf{Notes:}  
    \begin{itemize}
        \setlength\itemsep{0em}
        \item \textit{3B original} : Pretrained Qwen2.5-VL-3B-Instruct without fine-tuning.
        \item \textit{3B mlp} : Fine-tunes only the multimodal fusion MLP.
        \item \textit{3B mlp+vision} : Fine-tunes the fusion MLP and the vision encoder.
        \item \textit{3B mlp+llm} : Fine-tunes the fusion MLP and the LLM backbone.
        \item \textit{3B llm+mlp+vision} (VISTA): Fine-tunes all three components: vision encoder, fusion MLP, and LLM backbone.
    \end{itemize}
    }
\end{table}

 As shown in Table \ref{tab:ablations}, the pretrained model (\textit{3B original}) model performs the worst  across most metrics, underscoring the necessity of task-specific adaptation. Fine-tuning the MLP projector alone (\textit{3B mlp}) marginally improves structural alignment (BLEU, ROUGE), yet its semantic richness (METEOR, CIDEr) remains limited due to a frozen decoder incapable of adapting to downstream reasoning patterns.

The \textit{3B mlp+vision} variant introduces visual context but still retains a frozen language model. While visual features enhance structural grounding, the lack of LLM tuning hinders its capacity to interpret and articulate complex scene semantics. This gap is particularly evident in METEOR and CIDEr, both of which emphasize contextual fluency and consensus relevance.

Substantial performance gains are observed when the LLM backbone is unfrozen and tuned, as seen in \textit{3B mlp+llm}. Compared to the vision-tuning baseline, The model achieves notable improvements in CIDEr (+0.43) and METEOR (+0.04), suggesting that decoder tuning plays a pivotal role in aligning outputs with the narrative structure and reasoning patterns embedded in expert references. This observation is consistent with the findings in recent instruction-tuned LLM literature, highlighting decoder adaptation is central to cross-domain generalization.

Finally, our VISTA (\textit{3B llm+mlp+vision}) achieves the highest scores across nearly all metrics, demonstrating that joint optimizing vision, fusion, and language components yields synergistic improvements. The model leverage grounded visual cues, adaptive semantic decoding, and a well-aligned cross-modal embedding space. Its superior CIDEr and BLEU-4 scores suggest not only enhanced fluency but also precise alignment with human-level multi-step descriptions.


Overall, the results demonstrate that compact VLMs, when equipped with rich multi-agent supervision, can rival or even surpass the performance of larger single-agent models, offering a scalable, cost-efficient, and interpretable solution for video-based traffic risk assessment.

These findings further support the emerging perspective in multimodal learning that the quality of supervision data, prompt design, and alignment mechanisms can be more crtical than sheer model size, particularly for deploying lightweight VLMs for real-time applications.

\section{Conclusions}

This work introduces a scalable, modular framework for traffic scene understanding and safety risk assessment by leveraging multiple distinct VLM agents with structured prompting for knowledge distillation. Specifically, we orchestrate two complementary VLMs: GPT-4o for semantic scene interpretation and o3-mini for safety-centric reasoning, to generate high-quality, structured annotations from low-resolution traffic videos. These annotations, enriched through CoT prompting, serve as effective knowledge-enriched labels for fine-tuning a compact student model.

The resulting VISTA, a lightweight 3B VLM tailored for understanding traffic camera footage, delivers strong performance across common captioning metrics despite its compact size. These results highlight the crtical role of prompt design, collaborative agent supervision, and domain-specific grounding in developing efficient and interpretable VLMs for domain-targeted applications.

VISTA’s compactness and effectiveness make it well-suited for real-time deployment on edge devices within  existing transportation infrastructure. By augmenting  the analytical capability of legacy traffic cameras without hardware upgrades, our framework offers a cost-effective solution for enhancing incident detection and roadway safety at scale.

To promote reproducibility and foster community-driven adaptation, we publicly released both the model and the training pipeline . Practitioners can fine-tune VISTA on region-specific data to enable localized intelligence across diverse deployment contexts. This work lays the groundwork for future research in multi-agent prompting, structured knowledge distillation, and lightweight multimodal AI for intelligent transportation systems.

\section*{Acknowledgment}
This research was supported by the U.S. Department of Transportation, Office of the Assistant Secretary for Research and Technology (OST-R), University Transportation Centers Program, through the Center for Regional and Rural Connected Communities (CR2C2) under Grant No. 69A3552348304.

\section*{Author Contributions}
Conception and design: J.J.Y. and Y.Y.; data processing: Y.Y. and N.X.; analysis and interpretation of results: Y.Y., N.X. and J.J.Y.; draft manuscript preparation: Y.Y., N.X.; review and editing: J.J.Y.; visualization, Y.Y.; supervision, J.J.Y.; project administration, J.J.Y.; funding acquisition, J.J.Y. All authors have read and agreed to the published version of the manuscript.


\bibliographystyle{unsrt} 
\bibliography{references}  
\end{document}